\documentclass[conference]{IEEEtran}
\usepackage[numbers]{natbib}
\usepackage{multicol}
\usepackage[bookmarks=true]{hyperref}

\usepackage[ruled,vlined]{algorithm2e}
\usepackage{amsmath}
\usepackage{amsfonts}
\usepackage{authblk}
\usepackage{multirow}
\usepackage{epsfig} 
\usepackage{graphicx}
\usepackage{array}
\usepackage{hhline}
\usepackage{gensymb}
\usepackage{amssymb}
\newcolumntype{C}[1]{>{\centering\arraybackslash}p{#1}}

\usepackage{graphicx} % Required for inserting images
\usepackage[font=small,labelfont=bf]{caption}

\usepackage{color}
\usepackage{ulem}
\usepackage{caption}
\usepackage{subcaption}

\begin{document}

\title{Path Planning in 3D with Motion Primitives for Wind Energy-Harvesting Fixed-Wing Aircraft}

%\eric{Was a wordy title. Since the paper is mainly about the primitives I changed it.} 
% \seung{I checked it. Thanks}

%\title{3D Motion Primitives and SST Path Planning for Wind Energy-Harvesting Fixed-Wing Aircraft }

\author[1]{Seung-Keol Ryu}
\author[2]{Michael Moncton}
\author[3]{Han-Lim Choi}
\author[4]{Eric Frew}
\affil[1,3]{Department of Aerospace Engineering, Korea Advanced Institute of Science and Technology\\}
\affil[2,4]{Department of Aerospace Engineering Sciences, University of Colorado Boulder}

\maketitle

\begin{abstract}

In this work, a set of motion primitives is defined for use in an energy-aware motion planning problem. The motion primitives are defined as sequences of control inputs to a simplified four-DOF dynamics model and are used to replace the traditional continuous control space used in many sampling-based motion planners. The primitives are implemented in a Stable Sparse Rapidly Exploring Random Tree (SST) motion planner and compared to an identical planner using a continuous control space. The planner using primitives was found to run 11.0\% faster but yielded solution paths that were on average worse with higher variance. Also, the solution path travel time is improved by about 50\%. Using motion primitives for sampling spaces in SST can effectively reduce the run time of the algorithm, although at the cost of solution quality. 

\end{abstract}

\section{Introduction}

A standard limitation of many autonomous aerial robots and unmanned aircraft systems (UAS) is having a single energy source, batteries. Mounting bigger batteries is often not possible because of the increase in weight. However, other energy sources can be exploited, namely wind in the form of \textit{thermal updrafts}. Thermal updrafts push the aircraft upward. It will be beneficial in an energy consumption perspective. The exploitation of wind energy is more important for small UAS than for bigger UAS because small UAS don't have many spaces inside aircraft, and smaller spaces cause shorter endurance. 

%\eric{For such a short paper, the intro should get to the technical points quickly. These first two paragraphs, esp the very first, should be condensed to the main point or eliminated. Something like "Many autonomous aerial robots and uncrewed aircraft system (UAS) missions can benefit from extended endurance through wind energy extraction [citations]. Yada yada yada..."}

Gliders and many kinds of birds already utilize wind energy \cite{sachs2005minimum},\cite{richard2015upwind}, often in the form of thermal updrafts. When updrafts are exploited, energy extraction is called a \textit{static soaring}. If energy is extracted from a spatially-varying wind field, that is called a \textit{dynamic soaring}. This paper will only focus on \textit{static soaring}.

Considering the problem of extending the endurance of small UAS, a solution is proposed from the path planning perspective. Specifically, a balance must be struck between energy extraction and goal-oriented behavior. In other words, the aircraft must move towards the goal region from the start point while maximizing the usage of wind fields at the same time. This optimization is most useful in situations where fuel capacity is limited or the mission requires extended operation timescales. This research and methodology can help aircraft operate based not only on remaining fuel but also on ambient energy available in the environment \cite{akos2010thermal}.
% Write why the motion primitives are useful.

There are many methods of exploiting wind energy \cite{al2013wind}, \cite{edwards2010aut}, \cite{lolla2012path}, \cite{witt2008go} but the proposed algorithm uses \textit{Motion Primitives} which are presented by these papers \cite{de2019learning},\cite{paranjape2015motion},\cite{chakrabarty2013uav},\cite{babel2022online}. According to previous works, a fixed-wing aircraft is the best type of aircraft for exploiting wind energy. Because we decide to use a fixed-wing aircraft as a vehicle of path planner, we consider the dynamics of a non-holonomic vehicle. A non-holonomic vehicle can not go freely. That's why motion primitives are used as a method of our path planner. 

Motion primitives are defined as a pre-calculated set of motions that replace a traditional continuous action space. Aircraft motions are limited to the finite set of motion primitives, which form a representative subset of all possible movements of an aircraft. The virtue of motion primitives is that they can be used to explore the environment without much computation time and with little concern if the aircraft can follow the resulting path. Because the path planner-based motion primitives reduce the search space into a graph that connects with simple motions.

\section{Problem Statement}
Consider an obstacle-free environment $E \in \mathbb{R}^3$ with updraft currents to be navigated by an aircraft in an energy-efficient manner. Planning is done in a 4 degree-of-freedom 
% \eric{Why call this "reduced" since you have not described the "full" state? It is fine just to say this is the planning problem without justifying the state space} state space given by
\begin{gather}
\label{MP_dynamics} 
{\bf x} = [P_n,P_e,\chi,-h]^T
\end{gather}
where $P_n, P_e$ are the north (x) and east (y) components of the inertial position, $h$ is the aircraft height, and $\chi$, is the course angle.
Given a starting state $x_{start}$ and a goal region $X_{goal} \in E$, a planner seeks to find the most energy-efficient path from start to goal subject to the dynamic constraints given by
\begin{gather}
\label{simple_dynamics}
\dot{P_n} = u_{1}cos(\chi)cos(\gamma) + w_n \\
\dot{P_e} = u_{1}sin(\chi)cos(\gamma) + w_e \\
\dot{\chi} = u_{2} \\
\label{end_smp}
\dot{h} = u_{1}sin(\gamma) - w_d
\end{gather}
And the control inputs are given by,
\begin{equation}
U = \lbrace v_a,\dot{\chi},\gamma \rbrace = \lbrace u_1,u_2,u_3 \rbrace
\end{equation}

where $u_1$ is air relative airspeed, and $u_2$ is a time rate of $\chi$, and $u_3$ is a constant $\gamma$, flight path angle. The wind vector also has three elements $w_n,w_e,w_d$ (North, East, Down). 

\section{Path Planning Algorithm}

The path planner uses a Stable Sparse Rapidly Exploring Random Tree (SST) \cite{li2016asymptotically} algorithm using a set of motion primitives that simplify the action space. SST is well suited to the proposed problem because it eliminates the need to solve a complex, nonlinear two-point boundary value problem during the \textit{Steer} phase of other similar algorithms.

% \seung{I have a question of SST using motion primitives. Did it use purely random selected sampler space? I thought that the motion primitives in sampler space of SST compare each other and choose the best nodes. I thought that your steps are .
% 1. The several motion primitives is selected in randomly selected space.
% 2. Compare the evaluation functions ( subsection C.).
% 3. Choose the best one and the other motion primitives are removed. 
% Please let me know if I am wrong if I am wrong. 
% Then, "optimal" means, optimal for the evaluation functions of subsection C or your cost functions?}

% \mike{The essential structure of the algorithm is as follows:
% 1. "randomly" select a node
% 2. "randomly" sample a control input (either by picking a primitive or a value from a continuous control distribution)
% 3. propagate dynamics from node using sampled control
% 4. if node is valid and locally optimal, keep it otherwise discard and go to (1)
% 5. if propagated node displaces an existing node, recursive prune the branch.}

\subsection{Stable Sparse Rapidly Exploring Random Tree (SST)}
To evaluate the efficacy of using motion primitives in a planning context, a Stable Sparse Rapidly Exploring Random Tree (SST) \cite{li2016asymptotically} planner using motion primitives was compared to an SST planner utilizing a continuous control space. To do this, in the Monte-Carlo propagation phase of the planner, the algorithm using primitives selects a primitive at random and propagates the system dynamics according to the control input selected. In the continuous planner, the control input is selected from a continuous uniform distribution with the same span as the primitive set. Although both planners will use the same set of dynamics and the same envelope of control inputs, comparing the two will determine if preselected input sequences provide the planner with an advantage either in computation speed or in the optimality of the solution path. The \textit{cost function} of the planner Eq(\ref{eq.costs}) optimizes for minimum energy trajectories to the goal region.

% \seung{I change the word "the cost function" to "the evaluation function". Because the evaluation functions has the two parts, which are the heuristics to go, and the cost to come. If we say "the cost function" in this context, the heuristics of evaluation function could be confused. }\\

% \mike{It should be cost function. the evaluation function was what you were using for the exhaustive searching method, which eric has since removed. It is possible that we should remove the heuristic component, as I do not think it is being implemented anywhere currently.}
% I agree. You used only "the cost function" of evaluation part, right? If you do that, I think "Heuristics" part should be erased.
% sounds good
% I am erasing those parts. If you are okay, please remove the comments in the problem statement.
\subsection{Motion Primitives}
%Motion primitives are fundamentally a predefined set of motions intelligently selected to explore the environment. These primitives are generated using the dynamics model Eq(\ref{simple_dynamics})-Eq(\ref{end_smp}).

Motion primitives are a subset of the control input sequences that are expected to yield useful behaviors. 
Four types of motion primitives are defined. Those are \textit{straight}, \textit{curves}, \textit{spirals} Eq(\ref{eq.str,crv,spr}), and \textit{spline curves} Eq(\ref{eq.spl}). The \textit{straight} motions have constant heading angles. The \textit{curves} have to change heading angles and flight path angles. The \textit{spirals} are similar to \textit{curves}, but allow the aircraft to perform a half, single, two, or three circles. The \textit{spline curves} have two different heading angles. In the mid of the path, the path is turned to another heading. So, the total number of motion primitives is 174, which is 10 from (\textit{straight}), 60 from (\textit{curves}), 80 from (\textit{spirals}), and 24 from (\textit{spline curves}).

Let ${\bf u}(t) = \left[u_1(t), u_2(t), u_3(t)\right]^T,\;t \in [0,T_s]$ be the control input vector at discrete time $t$ and let ${\bf u}(t_0:t_n)$ be the discrete control sequence from time $t_0$ to $t_n$ where $t_{i+1} - t_i = T_s/N \; \forall i$. \(N\) is a number of intermediate points in predefined period \(T_s\).

The motion primitives are defined by taking combinations of inputs from different discrete sets. All motion primitives consider the same set $U^1$ of UAS speed
\begin{equation}
U_1 = \left[10, 20\right] \textrm{ m/s}
\end{equation}
and the same set $U_3$ of flight path angles
\begin{equation}
U_3 = \left[-45^{\circ}, -15^{\circ}, 0^{\circ}, 15^{\circ}, 45^{\circ}\right].
\end{equation}

The set $MP^{str}(t_1:t_N)$ of \textit{Straight} motion primitives has constant speed and constant flight path angle over the entire primitive duration. 
For the second and third set of \textit{Straight,Curve,Spirals} primitives, define the set $U_2^{str}$, $U_2^{crv}$ , $U_2^{spr}$ of turn rates. The motion primitives of three kinds are defined as Eq(\ref{eq.str,crv,spr}).
\begin{gather}
U_2^{str} = 0 ^{\circ}/s \\
U_2^{crv} = \dfrac{1}{T_s} \left[-90, -60, -30, 30, 60, 90 \right] ^{\circ}/s.\\
U_2^{spr} = \dfrac{1}{T_s} \left[-1080, -720, -360, -180, 180, 360, 720,1080 \right] ^{\circ}/s
\end{gather}

The \textit{spline} curves are slightly different from the other three kinds of motion primitives. The turn rate, $U_2$, has two different values in a single motion primitive. So, in ${U_2}^{spl}$, two values are chosen, one for each half of the trajectory segment. Along \textit{spline} curves, there is no guided height change leading to steady flight in the absence of wind.

\begin{equation}
U_2^{spl} = \dfrac{1}{T_s} \left[-90 , -60, 60, 90 \right] ^ {\circ}/s, U_3^{spl} = 0 ^{\circ}
\end{equation}

The structures of motion primitives are all similar. With three variables, the motion primitives are defined as sets Eq(\ref{eq.str,crv,spr}-\ref{eq.spl}).

% \eric{It looks like Straight, Curve, and Spiral primitives are all basically the same: fix the speed, turn rate, and climb rate. If so, I dont think they need to be defined separately...}

\begin{equation}
\label{eq.str,crv,spr}
\begin{array}{ll}
MP^{j}(t_1:t_N),& j = \lbrace str \rbrace ,\lbrace crv \rbrace,\lbrace spr \rbrace \\  = & \lbrace \left[u_1(t) \in U_1, u_2(t) \in U_2^{j}, u_3(t) \in U_3\right]^T | \\ 
& u_1(t_1) = u_1(t_2) = \cdots = u_1(t_N), \\ 
& u_2^{j}(t_1) = u_2^{j}(t_2) = \cdots = u_2^{j}(t_N), \\ 
& u_3(t_1) = u_3(t_2) = \cdots = u_3(t_N) \rbrace.
\end{array}
\end{equation}
\begin{equation}
\begin{array}{ll}
\label{eq.spl}
MP^{spl} =  & \lbrace \left[u_1(t) \in U_1, u_2^{1,2}(t) \in U_2^{spl}, u_3(t) \in U_3^{spl}\right]^T | \\ 
& u_1(t_1) = \cdots = u_1(t_N), u_2^{1} \neq u_2^{2}\\ 
& u_2^{1} = u_2(t_1) = \cdots = u_2(t_{N/2})\\
& u_2^{2} = u_2(t_{N/2+1}) = \cdots = u_2(t_N) \\
& u_3(t_1) = u_3(t_2) = \cdots = u_3(t_N) \rbrace.
\end{array}
\end{equation}

A trajectory is a series of motion primitives and states resulting from using these motion primitives. And the size of the trajectory is based on the depth of the final nodes of each trajectory. A trajectory is therefore defined as 
$Traj = \lbrace MP_{1},MP_{2},\cdots,MP_{Depth} \rbrace$. Fig. \ref{fig. mps} shows the complete set of primitives. The \textit{up} plot is a 3D view of motion primitives, and the \textit{down} plot is a bird-eye view of these.

\begin{center}
\includegraphics[width = 6cm]{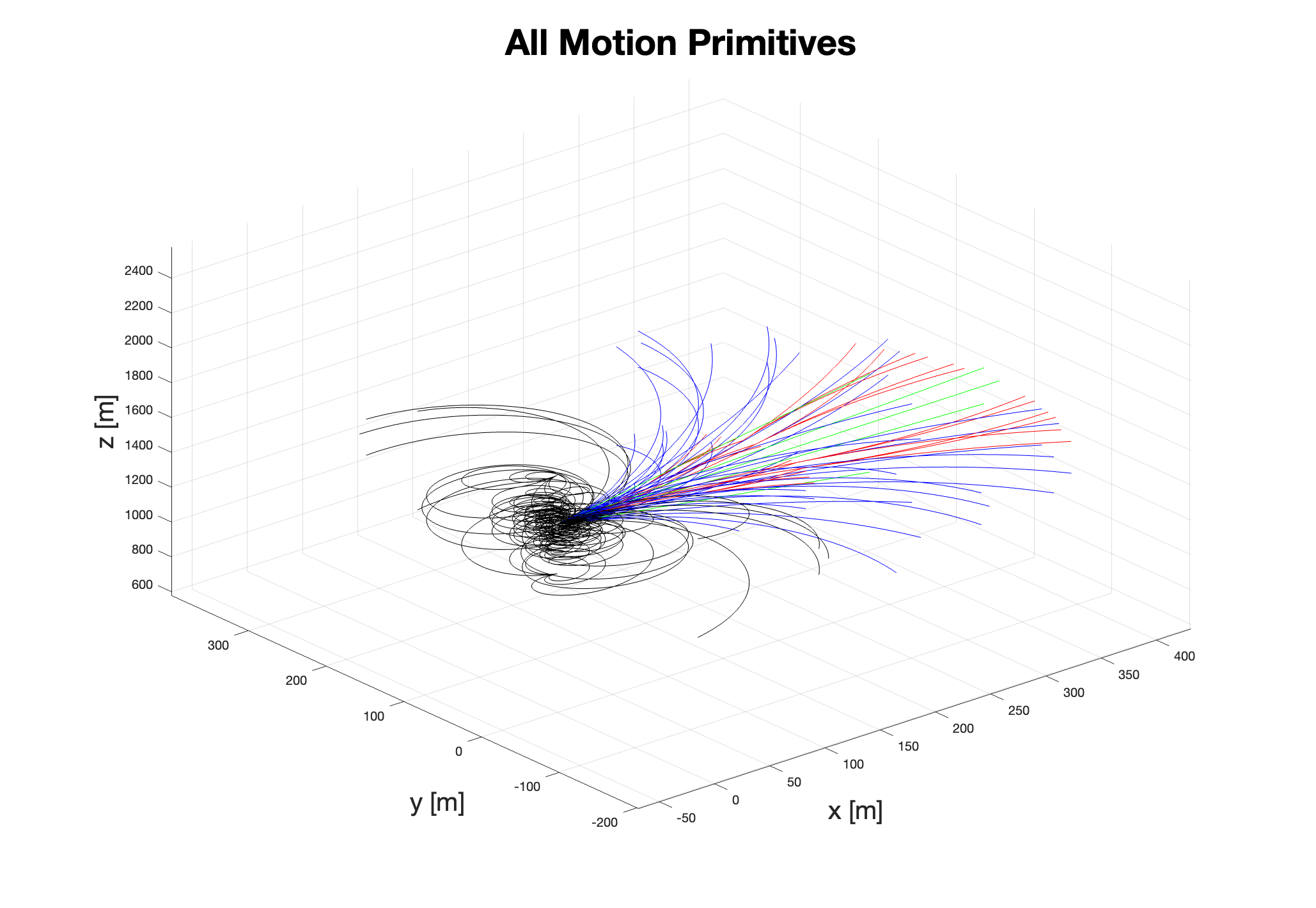}
\includegraphics[width = 6cm]{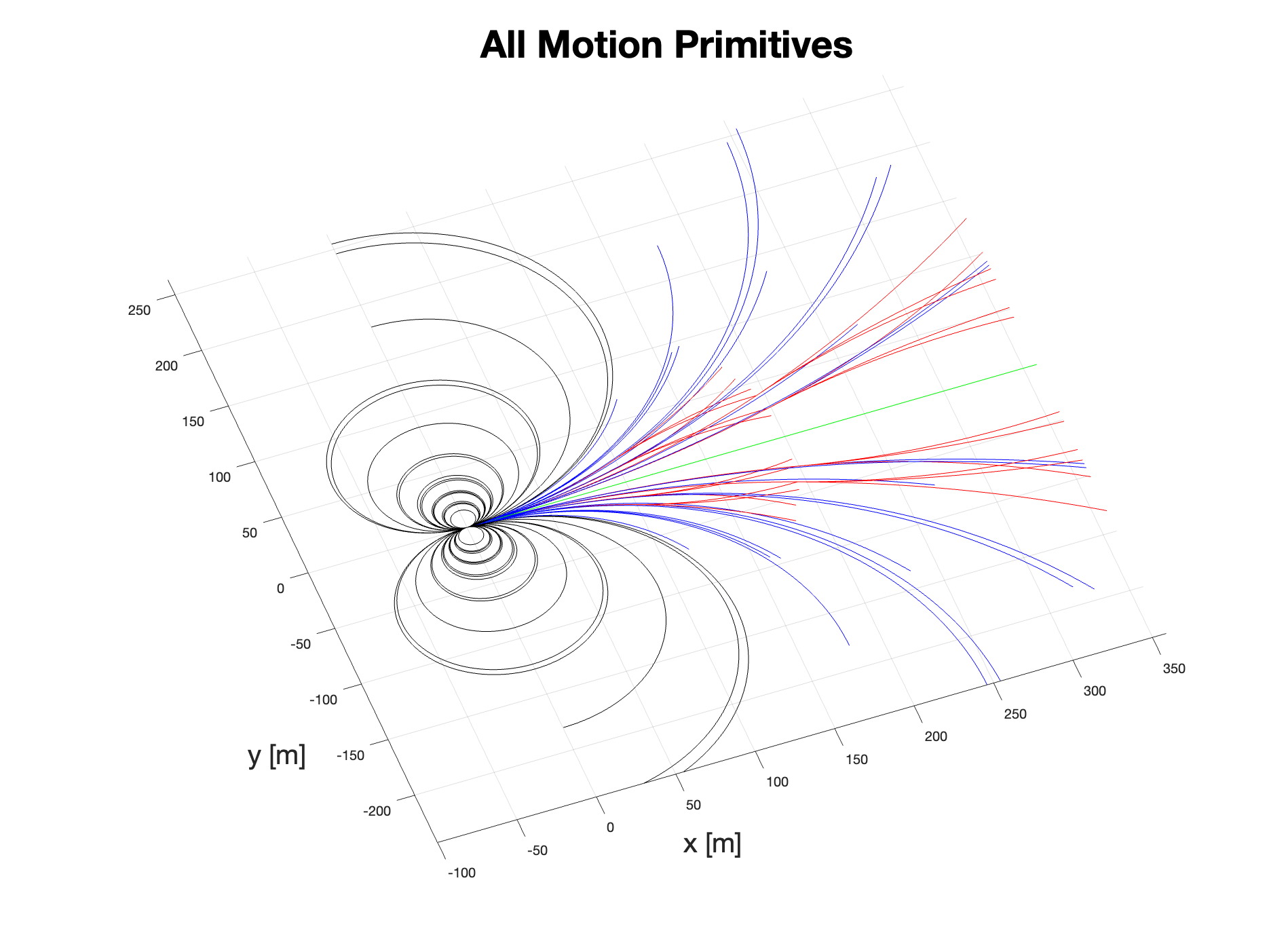}
\captionof{figure}{Expanding all elements of motion primitives from the origin. There are straight lines (\textit{green}), curves (\textit{blue}), spirals (\textit{black}), spline curves (\textit{red}) }
\label{fig.mps}
\end{center}

\subsection{Cost Function} \label{eval.funcs}

The purpose of using four kinds of motion primitives is to make a path planner explore quickly and effectively. However, it is computationally intractable to exhaustively explore all combinations of motion primitives. Therefore, a cost function Eq(\ref{eq.cost_func}) is necessary to pick the best motion primitives to generate further trajectories. The cost function is the energy consumption to come from the start node Eq(\ref{eq.costs}).
\begin{equation}
\label{eq.cost_func}
f(X_i) = c(X_{i},X_{start})
\end{equation}
The cost function Eq(\ref{eq.costs} is taken from \cite{chakrabarty2011energy}. The energy cost is the specific energy rate. The energy cost of each path, Eq(\ref{eq.costs}), has a potential energy term($ g\dot{h}_{start->i},h_{start->i}$: the height difference between the current node and the start node) term, a kinetic energy term($ v_a \dot{v_a}$). And the last term ($ \dot{e_f}_{start->i}$) is an internal fuel consumption term, Eq(\ref{eq.asmp}-\ref{eq.itn_fuel}). The kinetic energy term is neglected because, along the path between each node, the aircraft keeps $v_a$, the air relative velocity, constant. The gross velocity, $v_g$, is the summation of $v_a$ and wind velocity.
\begin{gather}
\label{eq.spc}
\frac{energy}{m}=e_{i}= gh_{start->i} + 0.5 (v_a)^2 + {e_f}_{start->i} \\
\label{eq.costs}
\dot{e_{i}} = g\dot{h}_{start->i} + v_a \dot{v_a} + \dot{e_f}_{start->i} \\
c(X_{i},X_{start}) = -\dot{e_{i}}/v_{g}
\end{gather}

Unlike the other terms, the internal fuel consumption term is not easily derived. Because according to \cite{chakrabarty2011energy}, the internal fuel consumption term Eq(\ref{eq.itn_fuel}), is a function of \textit{thrust} and $v_a$. But the \textit{thrust} is not easily derived. 

That's why two assumptions are used for the problem. The first assumption is that $v_a$ and $\gamma$, a flight path angle, are constant Eq(\ref{eq.asmp}). $\eta_{ec}$, $\eta_{p}$ are the energy efficiency constants of the vehicle we have used. The $\eta_{ec}$ is the net efficiency of conversion from energy source to shaft and the $\eta_{p}$ is the propeller efficiency.  We assume $\eta_{ec}$, $\eta_{p}$ as 0.8, 1 respectively.
\begin{gather}
\label{eq.asmp}
\dot{v_a} = 0 , \dot{\gamma} = 0 \\
\label{eq.itn_fuel}
\dot{e_f} = -\frac{T*v_a}{mg*\eta_{ec}*\eta_{p}}
\end{gather}
The second assumption is using simple polynomial equations of \textit{lift} and \textit{drag} forces Eq(\ref{eq.lft_drg}. The \textit{lift} and \textit{drag} are defined as,
\begin{gather}
q = 0.5 * {v_{a}}^2 * \rho(h) * S \\
\label{eq.lft_drg} 
L = q * C_{L_{0}},D = q * (C_{D_{0}} + {C_{L_{0}}}^2/(\pi*AR*e))
\end{gather}
The constants $C_{D_{0}},C_{L_{0}},AR,e$ are from the aircraft, \textit{Tempest} \cite{roadman2012mission}, which is a small fixed wing UAS.

The 3D point mass model from \cite{beard2012small} depicts the relationship between \textit{thrust}, \textit{lift}, and \textit{drag}. Each component of \textit{thrust} is related to \textit{lift}, and \textit{drag}, respectively. These assumptions, Eq(\ref{eq.asmp}), Eq(\ref{eq.lft_drg}), make the 3D point mass model \cite{beard2012small} of aircraft more easily derived. The simple model is a set of equations of $v_a$. According to Eq(\ref{eq.T_D}-\ref{eq.T_L}), the \textit{thrust} is a function of \textit{lift}, and \textit{drag}, which are modeled by the simplified polynomial equations. So the Eq(\ref{eq.va}) is for deriving ${T_x}$, x-direction of thrust,
\begin{multline} \label{eq.va}
\dot{v_a} = \frac{-D+T_x}{m}\ - gsin(\gamma) \\
- \left [cos(\chi)cos(\gamma),sin(\chi)cos(\gamma),sin(\gamma)\right] \frac{dW^E_E}{dt}\
\end{multline}
and Eq(\ref{eq.gam}) is for deriving ${T_z}$, z-direction of thrust. And the inertial wind velocity by inertial frame is $W_E^E$.
\begin{multline} \label{eq.gam}
v_a\dot{\gamma} = \frac{L-T_z}{m}\cos(\phi) - gcos(\gamma) \\ - \left [cos(\chi)sin(\gamma)\; sin(\chi)sin(\gamma)\; 
cos(\gamma)\right]\frac{dW^E_E}{dt}\
\end{multline}

The wind velocity of environments is constant, Eq(\ref{eq.Wind_constant}), except passing the thermals and using Eq(\ref{eq. asmp}) makes the right-hand side of equations Eq(\ref{eq.va}-\ref{eq.gam}) zero. 
\begin{gather}
\label{eq.Wind_constant}
 \frac{dW^E_E}{dt}\ = 0 \\
 \label{eq.T_D}
 T_x = D + mgsin(\gamma)\\
 \label{eq.T_L}
 T_z = L - \frac{mgcos(\gamma)}{cos(\phi)} 
\end{gather}

One of the assumptions of the 3D point mass model is that there are no side forces. In other words, the $T_y$ is zero allowing \textit{thrust} to be fully constrained. \textit{Thrust} is required for calculating the internal fuel consumption and if that is calculated, the costs Eq(\ref{eq.costs}) can be calculated at the same time.

\section{Results}
Both algorithms were run 30 times for 20 seconds. Of all of the runs, the continuous planner failed to find a solution path 2 times (6.7\%) and the primitive planner failed to find a solution 3 times (10\%). In the remaining runs where a solution path was found, the two algorithms varied significantly in terms of computational performance, and the output graphs that were solved for. The primitive-based planner ran 11\% faster than the continuous planner while maintaining the same number of active nodes in the graph.

\begin{center}
\includegraphics[width = 6cm]{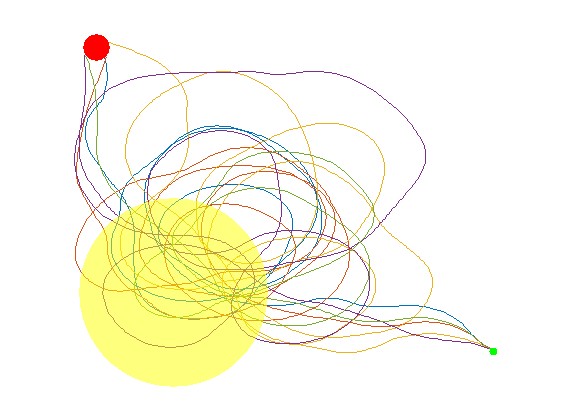}
\captionof{figure}{A set of five example solution trajectories from an initial state (lime) to goal region (red). A thermal is presented in yellow.}
\end{center}

\noindent
\begin{center}
\includegraphics[width = 7cm]{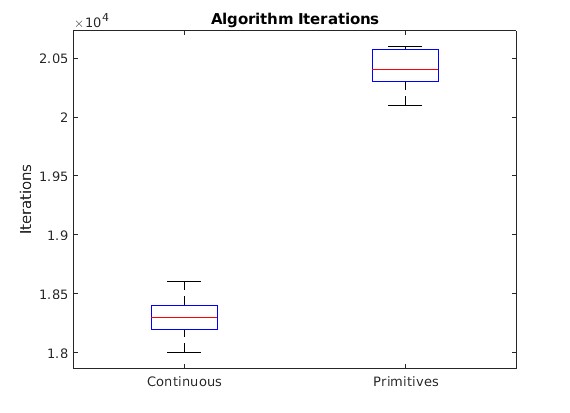}\\
\captionof{figure}{Number of algorithm iterations per 20 seconds}
\label{RunTime}
\end{center}

However, the continuous planner outperformed the primitive-based planner on average in terms of quality of solution path cost ($3.51e5$ $\pm$ $2.97e4$ vs $7.32e5$ $\pm$ $1.39e5$).

\noindent
\begin{center}
\includegraphics[width = 7cm]{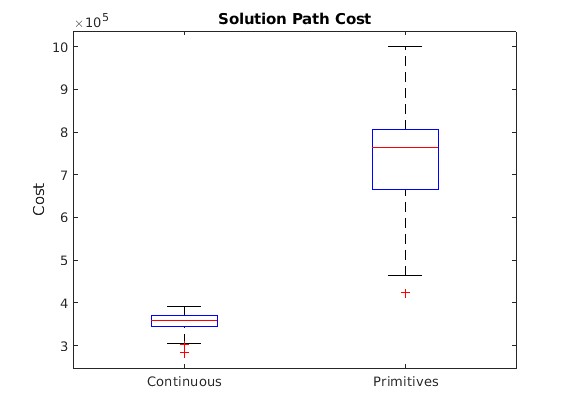}
\captionof{figure}{Soltion path cost using energy consumption heuristic}
\label{PathCost}
\end{center}

Although the planner was optimizing for path cost alone, an effect of using primitives was a reduction in flight time from start to goal. The primitive-based planner found paths that took almost half as long to fly as the continuous planner despite flight time not being an optimization criterion.  

\begin{center}
\includegraphics[width = 7cm]{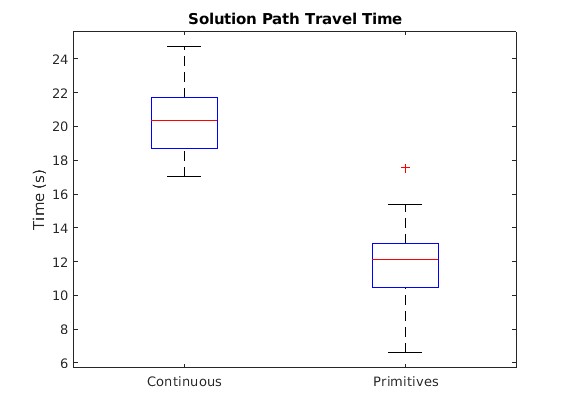}\\
\captionof{figure}{Flight time of solution paths}
\label{ActiveNodes}
\end{center}

\section{Discussions}
The results from multiple independent runs of both algorithms demonstrate the significant differences between both approaches as well as advantages and disadvantages. Planning using a continuous input space allows the planner to find a more optimal path reliably, possibly because the true optimal path lies between primitives and is unavailable for sampling by the primitive-based planner. However, simplifying the problem by using a finite set of motion primitives allows the algorithm to run faster while still saturating the environment with active nodes. The significant reduction in path travel time, however, is an unexpected result, as there is no direct reason either planner should optimize for travel time. Further analysis is required to determine if a heuristic consisting of a combination of travel time and energy would allow the continuous planner to reduce path travel time while maintaining a lower solution path cost.

\bibliographystyle{IEEEtran}
\bibliography{references}

\end{document}